\documentclass[12pt]{article}

\usepackage{amsmath,amssymb,amsfonts}
\usepackage{array}
\usepackage{algorithm2e}
\usepackage{tabularx}
\usepackage{subcaption}
\usepackage{graphicx}
\usepackage[biblabel]{cite}
\usepackage[colorlinks=true, linkcolor=blue, citecolor=blue, urlcolor=blue]{hyperref}

\usepackage[a4paper,margin=25mm]{geometry}

\begin{document}

\title{A Symmetry-Integrated Approach to Surface Code Decoding}

\author{
    Hoshitaro Ohnishi\thanks{Computer Science Program, Graduate School of Science and Technology, Meiji University, 1-1-1 Higashimita, Tama-ku, Kawasaki, Kanagawa 214-8571, Japan} 
    \and
    Hideo Mukai\thanks{Computer Science Program, Graduate School of Science and Technology, Meiji University, 1-1-1 Higashimita, Tama-ku, Kawasaki, Kanagawa 214-8571, Japan; Department of Computer Science, School of Science and Technology, Meiji University, 1-1-1 Higashimita, Tama-ku, Kawasaki, Kanagawa 214-8571, Japan. Corresponding Author, e-mail: mukai@meiji.ac.jp}
}

\date{}

\maketitle

\begin{abstract}
Quantum error correction, which utilizes logical qubits that are encoded as redundant multiple physical qubits to find and correct errors in physical qubits, is indispensable for practical quantum computing. Surface code is considered to be a promising encoding method with a high error threshold that is defined by stabilizer generators. However, previous methods have suffered from the problem that the decoder acquires solely the error probability distribution because of the non-uniqueness of correct prediction obtained from the input. To circumvent this problem, we propose a technique to reoptimize the decoder model by approximating syndrome measurements with a continuous function that is mathematically interpolated by neural network. We evaluated the improvement in accuracy of a multilayer perceptron based decoder for code distances of 5 and 7 as well as for decoders based on convolutional and recurrent neural networks and transformers for a code distance of 5. In all cases, the reoptimized decoder gave better accuracy than the original models, demonstrating the universal effectiveness of the proposed method that is independent of code distance or network architecture. These results suggest that re-framing the problem of surface code decoding into a regression problem that can be tackled by deep learning is a useful strategy.

\end{abstract}

\section{Introduction}
Quantum computation is an information processing technique that exploits quantum mechanical properties such as superposition and entanglement and it has the potential to exceed the capabilities of classical computation in specific contexts\cite{shor1994,broadbent2020,zhou2020,suzuki2024,kim2024,amin2018,he2018,ibe2022,ohgoe2024}. However, quantum computations require quantum error correction (QEC) because of the sensitivity of qubits to external noise namely in the form of heat or electromagnetic waves \cite{weber2024}.

QEC utilizes logical qubits encoded with redundant multiple physical qubits to find and correct errors in physical qubits. The stabilizer code is defined by a stabilizer generator and errors are detected by performing a syndrome measurement\cite{gottesman1997}. Whereas classical computations using bits are affected by only bit-flip errors, quantum computations using qubits are affected by various other errors such as phase-flip errors, making QEC a more complicated problem than classical error correction.

Surface code is a promising paradigm that possess a high error threshold and good implementability, toric code being one such implementation\cite{kitaev2003}. Moreover, minimum-weight perfect matching (MWPM) is a prominent classical decoding algorithm of surface code\cite{raussendorf2007,iolius2023}. Although MWPM can attain high decoding accuracy, its computational time is polynomial with code distance, thus presenting a scalability issue. Deep learning decoding implementations, such as multilayer perceptrons (MLP), convolutional neural networks (CNN), graph neural networks (GNN), and transformers, have been extensively studied and have been found to learn appropriate noise models \cite{Krastanov2017,Wang2022,Wu2024,Lange2023,FjelddahlBengtsson2024,Wang2023,fitzek2020,varsamopoulos2017,varbanov2025,hu2025}.

Decoders with deep architectures suffer from the essential aspects of the quantum error correction problem. In particular, maximum likelihood decoding of typical stabilizer code, as well as toric code, is an NP-hard problem\cite{prxquantum2023}. This is because the input error patterns of each syndrome, which are the correct answers, do not correspond uniquely to the deep structures in the decoder training. Such decoders have thus been optimized in training to output the most frequent pattern of the errors as the error corresponding to the syndrome within the prepared dataset. For example, the decoder does not consider whether the output correction operator resolves asymmetries of the coded state or not. This problem also causes an exponential increase in the size of the training dataset needed to achieve high accuracy when the deep architecture decoder learns a large code distance.

To address these issues, we redefined the syndrome measurement as a mathematically equivalent function and devised a reoptimization method for the decoder with an approximation model of  neural network for the extrapolated continuous function. This approach makes it possible to consider the symmetry unable to be accomplished by the original decoder, improving the accuracy of the decoder obtained from the same dataset.

We experimented with the proposed method in an MLP decoder for code distances of $5$ or $7$ and compared the decoding accuracies before and after reoptimization. We also experimented with our method in the decoders with CNN, RNN, and transformer architecture for code distances of $5$. Furthermore, we evaluated the size of the dataset needed for a non-reoptimized decoder to achieve the same accuracy as that of the MLP decoder reoptimized with our method for a code distance of $5$ in order to see the extent of the dataset reduction enabled by the reoptimization. Finally, we investigated the effectiveness of the reoptimization when the training dataset was one sampled from a polarized noise model with specific errors.

The results showed that the reoptimization improved accuracy in all cases. Notably, the transformer decoder for a code distance of $5$ had a $60.7\%$ improvement in accuracy. The results also revealed that, without reoptimization, a 5-fold larger dataset would be needed to in order to reach the level of the accuracy of the reoptimized MLP decoder for a code distance of $5$; this demonstrates the effect that reoptimization with our current method has reduced the required size of the training dataset. 

\section{Method} 
\label{2}
\subsection{Problem Setting} 
\label{2.1}
When the code distance is set as $L$ in toric code, $L^2$ $X$ stabilizer operators and $Z$ stabilizer operators have to be placed on a two-dimensional torus. In the current study, the vector $s \in \{0,1\}^{2L^2}$ was designed to contain values corresponding to the stabilizer operator measuring eigenvalues of $1$ if the state is error free and $0$ otherwise. For a physical qubit, $2L^2$ qubits must be placed on the grid of a two-dimensional torus. Here, we prepared two $2L^2$ vectors, each containing bit-flip errors as well as phase-flip errors for each physical qubit and concatenated them into $4L^2$ dimensional vector $e \in \{0,1\}^{4L^2}$”.

We assumed that the noise models followed a discrete uniform distribution and a category distribution. For the discrete uniform distribution, the probability of generation of every error is $\frac{p}{3}$. Regarding the category distribution, the probability of a bit-flip or phase-flip error is $\frac{p}{\eta+2}$ and that of an amplitude-phase error is $\frac{\eta p}{\eta+2}$. Here, $p$ is the error probability of any kind occurring in a single qubit.

Measurements of the decoder accuracy of the above model were carried out $10,000$ times for each of the following rates: from $0.1\%$ to $5\%$ in steps of $0.1\%$. Data restoration was considered as successful when syndromes were not detected from the coded states after error correction and logical operators were not acting on the code space. We considered that the states were logical errors when either of these two conditions were violated.

\subsection{Approximation of the continuous function interpolated the syndrome measurement} 
\label{2.2}
In syndrome measurements of toric code, eigenvalues are measured by $X$ and $Z$ stabilizer operators, which we will reformulate as mathematically equivalent functions. Here, a stabilizer operator acts on the surrounding up, down, the left and right physical qubits in the lattice. For the $X$ stabilizer, an eigenvalue of $-1$ is measured when a phase-flip error occurs in an odd number of qubits out of $4$ physical qubits; otherwise, the eigenvalue is $1$. To obtain a vector $s$ from the error vector $e$, each element of $s$ is calculated as the remainder of the sum of the elements of $e$ divided by $2$, which represents the error states of the $4$ physical qubits around the stabilizer operator.\par
Therefore, the function is:
\begin{equation}
    f:[0,1]^{4L^2} \to [0,1]^{2L^2},
\end{equation}
\begin{equation}
    f = (g_1,g_2, \dots ,g_{L^2},h_1,h_2, \dots ,h_{L^2}),
\end{equation}
\begin{equation}
    g_k = \frac{1-cos\pi x'}{2},
\end{equation}
\begin{equation}
    h_k = \frac{1-cos\pi x''}{2}.
\end{equation}
Here, to obtain the same output as the syndrome measurement as we formulated earlier, we let $x'$ and $x''$ be the sum of the elements of $e$ corresponding to $4$ physical qubits surrounding each stabilizer operator. 

To approximate the extended syndrome measurement function $f$ with a neural network, we sampled pairs of the vector $e' \in [0,1]^{4L^2}$ and vector $s' \in [0,1]^{2L^2}$ from a uniform distribution and used them as a training dataset. The approximated function $f$ is continuous in a compact subspace of Euclidean space, giving it a theoretically guaranteed arbitrary level of accuracy by the universal approximation theorem \cite{hornik1991}. In particular, $f$ was approximated with a neural network having the settings in Table~\ref{tab:ApproxNN}. 
\begin{table}[htbp]
\centering
\renewcommand{\arraystretch}{1.0}
\begin{tabular}{@{} c|c @{}}
Parameters & Values \\
\cline{1-2}
Input dimension & $4L^2$ \\
Output dimension & $2L^2$ \\
Number of hidden layers & $1$ \\
Hidden layer dimension & $hidden scale \times \text{input dimension}$ \\
Activate function & SeLU \\
Output layer activate function & Sigmoid \\
Number of training data & $10^6$ ($10^7$ when $L=7$) \\
Number of test data & $10^4$ \\
Batch size & $2^9$ ($2^{11}$ when $L=7$) \\
Epoch & 30 \\
Learning rate & $0.00001$ \\
$hidden scale$ & $1000$ ($750$ when $L=7$) \\
Loss function & MSE \\
Optimizer & AdamW \\
\end{tabular}
\caption{Neural network training settings for approximating a continuous function. $L$ indicates the code distance.}
\label{tab:ApproxNN}
\end{table}

Here, $hiddenscale$ is a hyperparameter to be set in training. Table~\ref{tab:ApproxNN} shows the number of training data and the other hyperparameters. 

\subsection{Training the Decoder} 
\label{2.3}
First, we trained the MLP decoder to predict the error vector $e$ from the corresponding vectors $s$ that represent the results of the syndrome measurement. The parameters used to train the MLP are shown in Table~\ref{tab:MLPset}.
\begin{table}[htbp]
\centering
\renewcommand{\arraystretch}{1.0}
\begin{tabularx}{\linewidth}{@{} c|c c @{}}
Parameters & First Training & Reoptimization \\
\cline{1-3}
Input dimension & $4L^2$ & -\\
Output dimension & $2L^2$ & - \\
Number of hidden layers & $18$ & - \\
Hidden layer dimension & $hidden scale \times \text{input dimension}$ & - \\
Activate function & SeLU & -\\
Output layer activate function & Sigmoid & - \\
Number of training data & $2 \times 10^6$ ($4 \times 10^7$ when $L=7$) & -\\
Number of test data & $10^5$ ($2 \times 10^5$ when $L=7$) & -\\
Batch size & $500$ & $200$ ($400$ when $L=7$) \\
Epoch & 55 & 75\\
Learning rate & $0.0005$ & $3 \times 10^{-8}$ \\
$hidden scale$ & $8$ & - \\
Loss function & BCE & BCE \\
Optimizer & Adam & Adam \\
\end{tabularx}
\caption{Settings for training and renormalization of MLP decoder.}
\label{tab:MLPset}
\end{table}
Here again, we used $hiddenscale$ as a hyperparameter in the training. The number of data and the other hyperparameters are listed in Table~\ref{tab:MLPset}. Next, the weights in the trained MLP decoder were reoptimized using the neural-network-approximated continuous function, as explained in section 2.2. The dataset for the reoptimization was the same as in the training of the decoder. Here, we first supplied the trained decoder (hereafter referred to as $g$) with a vector $s$ retaining the syndrome measurement results. From the obtained results $e'$ and the error vector $e$ corresponding to $s$, we calculated $|e-e'|$ as the input to $g$. Loss functions for evaluating the differences between the results provided by g and the zero vector were estimated and optimized based on gradients. In the optimization, weight parameters of the decoder were updated by considering the symmetry of the encoded states. The input to the neural network $g$ that approximates the continuous function for  interpolating syndrome measurements was $|e-e'|$, which implies a decoding calculation based on the obtained error prediction.
\begin{algorithm}[H]
\caption{Re-Optimizing Decoder Algorithm}
\KwIn{syndrome $s$, error $e$}

Load the pretrained decoder model\;
Load the model approximating the function $f$ and fix the parameters\;

\For{epoch}{
    $e' \leftarrow \mathrm{decoder}(s)$\;
    $s' \leftarrow f(|e - e'|)$\;
    $loss \leftarrow \mathrm{Binary Cross Entropy Loss}(s', \boldsymbol{0})$\;
    Update decoder mode parameters\;
}
\end{algorithm}
Because a zero vector is output when the symmetry completely vanishes with no syndrome, the losses were estimated in terms of the difference from the zero vector. The hyperparameters used for training are shown in Table~\ref{tab:MLPset}. 

We also trained the decoder by using a CNN, RNN, or transformers to predict the error vector $e$ from the syndrome measurement result vector $s$ and compared their performances with that of the MLP-based decoder. The correct data $e$ used in the training were sampled from a discrete uniform distribution with $p=0.05$. Data and hyperparameters for each decoder are shown in Table~\ref{tab:ARCset}. The hyperparameters for reoptimization are described in Table~\ref{tab:ARCset_reop}. 
\begin{table}[htbp]
\centering
\renewcommand{\arraystretch}{1.0}
\begin{tabular}{@{} c|c c c @{}}
Parameters & CNN & RNN & transformers \\
\cline{1-4}
Number of training data & $2 \times 10^6$ & $2 \times 10^6$ & $2 \times 10^6$ \\
Number of test data & $10^5$ & $10^5$ & $10^5$ \\
Batch size & $500$ & $500$ & $500$ \\
Epoch & $55$ & $55$ & $55$ \\
Learning rate & $0.00005$ & $0.00005$ & $0.0005$ \\
Loss function & BCE & BCE & BCE \\
Optimizer & Adam & Adam & Adam \\
$hidden scale$ & - & $16$ & - \\
Number of layers & - & $10$ & - \\
Embedding dimension & - & - & $128$ \\
Number of heads & - & - & $8$ \\
Number of encoder layers & - & - & $4$ \\
\end{tabular}
\caption{Training settings for CNN, RNN, and transformer decoders for code distance $5$.}
\label{tab:ARCset}
\end{table}

\begin{table}[htbp]
\centering
\renewcommand{\arraystretch}{1.0}
\begin{tabular}{@{} c|c @{}}
Parameters & All Architectures \\
\cline{1-2}
Batch size & $200$ \\
Epoch & 75 \\
Learning rate & $3 \times 10^{-8}$ \\
Loss function & BCE \\
Optimizer & Adam \\
\end{tabular}
\caption{Settings for reoptimization of CNN, RNN, and transformers for a code distance of $5$. The parameters are the same for all architectures.}
\label{tab:ARCset_reop}
\end{table}

We employed ResNet18 \cite{he2016} as the CNN decoder. The RNN decoder was based on modules in PyTorch. The input and output dimensions were the same as in the MLP, and the number of dimensions of the hidden layer was set to $hiddenscale$ as well. The transformer decoder used the encoder block architecture \cite{vaswani2017}. 

\section{Results and discussion} 
\label{3}
\subsection{Approximation of the continuous-function-interpolated syndrome measurement} 
\label{3.1}
We evaluated the model for the continuous function approximation for a code distance $5$ or $7$ on $10000$ input data sampled from a uniform distribution. As evaluation indices, we employed cosine similarity, mean square error (MSE), and mean absolute error (MAE) (Table~\ref{tab:result_app}). The obtained function approximation had high accuracy and the loss function converged in training. 
\begin{table}[htbp]
\centering
\renewcommand{\arraystretch}{1.0}
\begin{tabular}{@{} c|c c @{}}
evaluation index& $L=5$ & $L=7$ \\
\cline{1-3}
cosine similarity & $0.99939$ & $0.99948$ \\
MSELoss & $0.00040$ & $0.00033$ \\
L1Loss & $0.01477$ & $0.01281$ \\
\end{tabular}
\caption{Results of evaluation of the ability of model using a continuous function approximation.}
\label{tab:result_app}
\end{table}

\subsection{Effect of Code Distance} 
\label{3.2}
\begin{figure}
\centering
\includegraphics[scale=0.5]{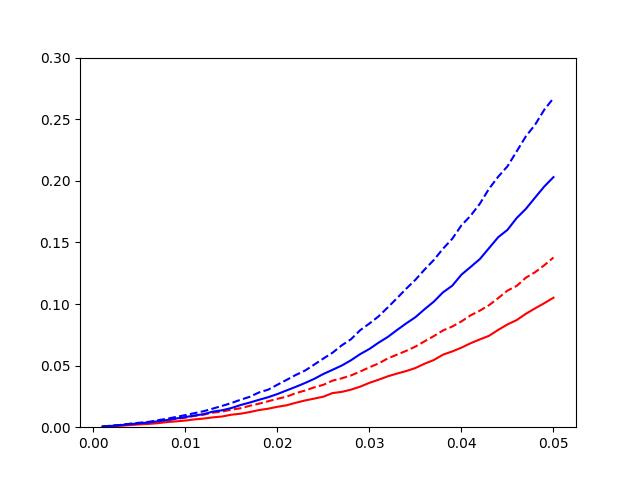}
\caption{Horizontal axis: error rate of physical qubits. Vertical axis: logical error rate of the decoder. Red: code distance $5$, blue: code distance $7$. Dotted line: Before reoptimization, Solid line: After reoptimization.}
\label{distane_MLP}
\end{figure}

First, we performed training until the loss converged. Here, we wanted to confirm that no room was left for improving the optimization method and to demonstrate the specific effects of the subsequent reoptimization after the first optimization. We repeated the experiments with code distances of $5$ or $7$ $20$ times each and measured the mean and standard deviation of the difference in the logical error rate before and after reoptimization (Fig.~\ref{distane_MLP}) .

\begin{figure}[htbp]
  \centering
  \begin{subfigure}[b]{0.45\linewidth}
    \centering
    \includegraphics[width=\linewidth]{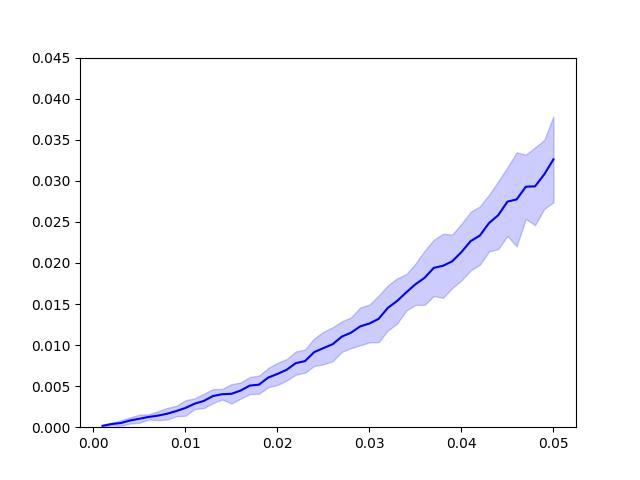}
 \caption{$L=5$}
    \label{fig:MLP_5}
  \end{subfigure}
  \hspace{0.05\linewidth}
  \begin{subfigure}[b]{0.45\linewidth}
    \centering
    \includegraphics[width=\linewidth]{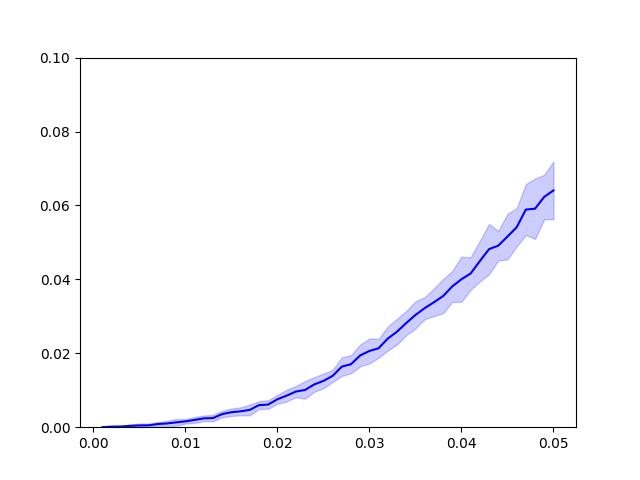}
\caption{$L=7$}
    \label{fig:MLP_7}
  \end{subfigure}
\caption{Logical error rate for $L=5,7$ before and after reoptimization by MLP decoder. Horizontal axis: error rate of physical qubit. Vertical axis: logical error rate of the decoder. Solid line: mean of difference in logical error rate before and after reoptimization. Blue shadow: Area within standard deviation.}
  \label{fig:MLP_distance}
\end{figure}

Decoder accuracy after training was higher for $L=5$ than for $L=7$ both before and after reoptimization. The dimensions of the input, output, and hidden layer grew quadratically with the code distance $L$. Compared with the case of $L=5$, the dimensions of each layer were $1.96$ fold larger in the case of $L=7$. Thus, the size of the dataset needed to obtain the same degree of accuracy increased faster than the dimensions of input or output. Next, we calculated the mean and standard deviation of the difference in logical error rate for code distances of $5$ and $7$ before and after reoptimization (Fig.~\ref{fig:MLP_5} and ~\ref{fig:MLP_7}).

The results show that the effect of the reoptimization on the standard deviation was independent of the code distance. The reoptimization effects depend quadratically on the error rate of the physical qubit.

\subsection{Comparison of Different Decoder Architectures} 
\label{3.3}
\begin{figure}
\centering
\includegraphics[scale=0.5]{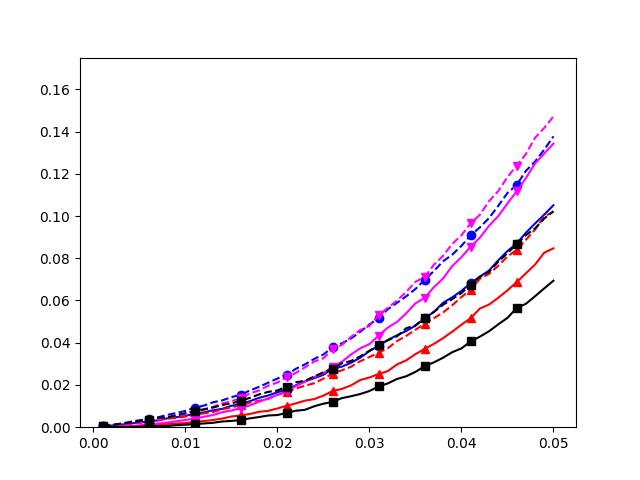}
\caption{Mean logical error rate of decoders before and after reoptimization. Horizontal axis: error rate of physical qubit, Vertical axis: mean of logical error rate of decoders. Dotted line: mean of logical error rate before reoptimization, solid line: after reoptimization. Red: MLP, blue: CNN, green: RNN, and cyan: transformer.}
\label{Arc}
\end{figure}

We compared the reoptimization effect of the MLP, CNN, RNN, and transformer decoders for a code distance of $5$. We used the data of the same size and calculated the means and standard deviation on the results of $20$ trainings (Fig.~\ref{Arc}). 

\begin{figure}[htbp]
  \centering
  \begin{subfigure}[b]{0.3\linewidth}
    \centering
    \includegraphics[width=\linewidth]{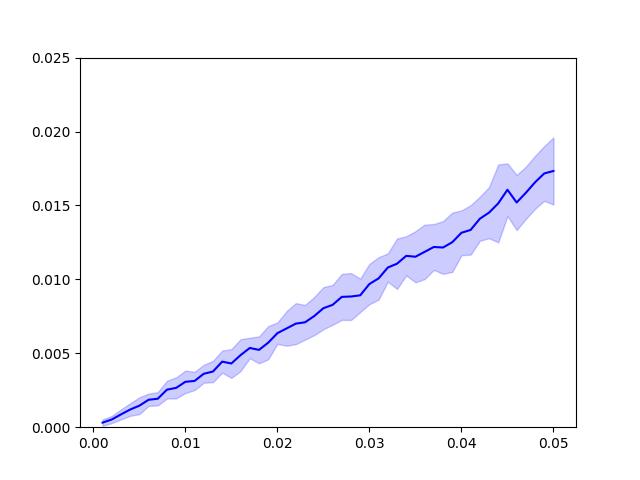}
 \caption{CNN}
    \label{fig:CNN}
  \end{subfigure}
  \hfill
  \begin{subfigure}[b]{0.3\linewidth}
    \centering
    \includegraphics[width=\linewidth]{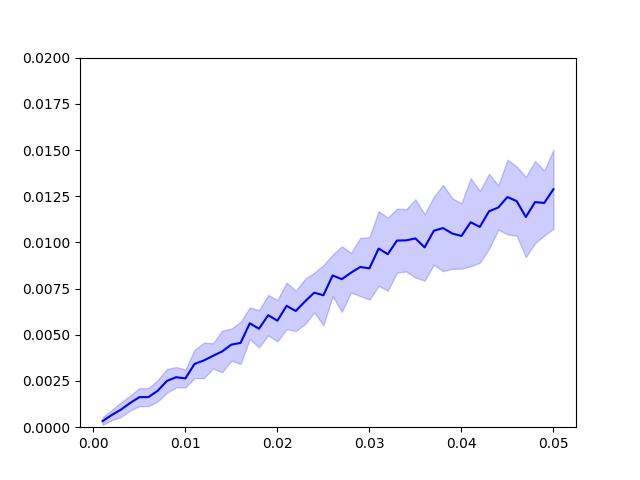}
 \caption{RNN}
    \label{fig:RNN}
  \end{subfigure}
  \hfill
  \begin{subfigure}[b]{0.3\linewidth}
    \centering
    \includegraphics[width=\linewidth]{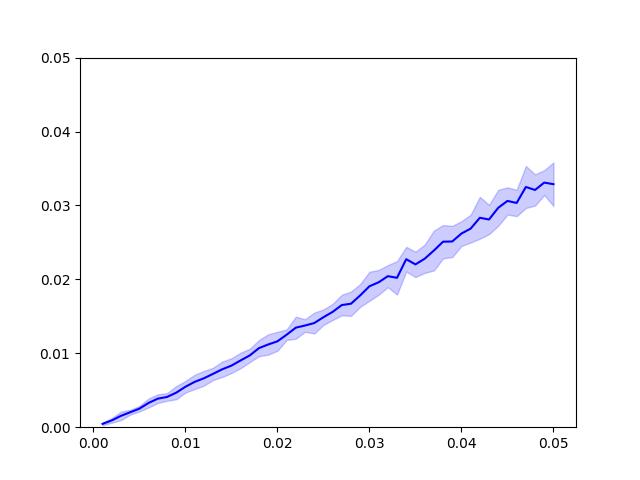}
 \caption{Transformers}
    \label{fig:TRF}
  \end{subfigure}
\caption{Means and standard deviations of logical error rates of CNN, RNN, and transformer-based decoders for a code distance of $5$. The legend in the figure is the same as in Fig.~\ref{fig:MLP_distance}.}
\label{fig:ARC}
\end{figure}

The transformer decoder outperformed the decoders based on the MLP, CNN, and RNN. Moreover, reoptimization had the smallest effect on the RNN decoder. The means and standard deviations of the logical error rates of the CNN-, RNN-, and transformer-based decoders are shown in Fig.~\ref{fig:CNN}, ~\ref{fig:RNN}, and ~\ref{fig:TRF} for a code distance of 5.

These results suggest that our reoptimization method works well with any of these architecture. The effect of reoptimization on the error rate of the physical qubits was linear in the CNN- and transformer-based decoders, while it was quadratic in the RNN decoder.

\subsection{Effect of the Number of Training Samples} 
\label{3.4}
\begin{figure}
\centering
\includegraphics[scale=0.5]{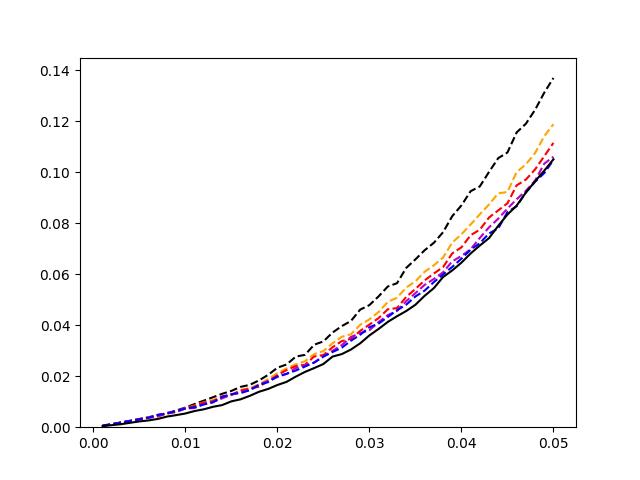}
\caption{Means of logical errors of the decoder trained on data from x1 to x5. Horizontal axis: error rate of physical qubit. Vertical axis: mean of logical error rate. Black: result after training with x1 data, green: x2, blue: x3, orange: x4, and magenta: x5. Black dotted line: mean of logical error rate after reoptimization with x1 data.}
\label{Data_scale}
\end{figure}

Next, we compared the performances of the decoders without reoptimization when they were trained on datasets of different sizes (from two-time and five-times the size of the dataset used to train the reoptimized decoders) The code distance was $5$ in all of the experiments and the mean of the logical errors was calculated for $10$ trials for each decoder. The architecture of the decoders was MLP. We compared the results with those of the decoder reoptimized with the original number of data (Fig.~\ref{Data_scale}). 

An inspection of Fig.~\ref{Data_scale} reveals that the decoder reoptimized with our method achieved the same accuracy as the unoptimized decoder but with an $80$\% reduction in training data. 

\subsection{Comparison under Different Noise Bias Conditions} 
\label{3.5}
\begin{figure}
\centering
\includegraphics[scale=0.5]{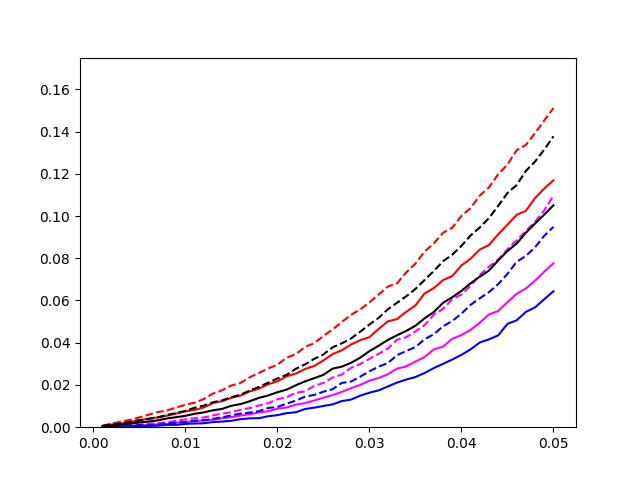}
\caption{Effect of the skewness of noise model. Horizontal axis: error rate of physical qubit. Vertical axis: mean of logical error rate. Red: $\eta=0.5$, magenta: $\eta=3$, blue: $\eta=5$. Black: $\eta=1$ without skewness. Dotted line: before reoptimization. Solid  line: after reoptmization.}
\label{Biased}
\end{figure}

Lastly, we show the results of sampling the vector $e$ from the category distribution defined in section 2.1 for use as correct data in the training of the decoder. Here, $\eta$ is the parameter of the category distribution. The architecture was MLP and the code distance was $5$. We also performed the experiments with $\eta = 0.5, 3,$ and $5$. The decoding accuracies are shown for each value of $\eta$ in Fig.~\ref{Biased}, where each plot is an average obtained over 10 trails.

The decoding accuracy became better and the reoptimization had a larger effect as the correlation between bit-flip errors and phase-flip errors increased. This is because the problem to be solved by the model becomes simpler as the value of $\eta$ increases, i.e., as the tendencies of the bit-flip error and phase-flip error become more similar. The same holds for reoptimization. These results suggest that our reoptimization method is effective in the training the decoder in the noise model that accompanies amplitude-phase errors with relatively high probability.

\section{Conclusions} 
\label{4}
We proposed a reoptimization method to improve the accuracy of toric-code decoders with deep architectures. We demonstrated the reproducibility of reoptimizing MLP decoders for code distances of $5$ and $7$. For physical qubits with a $5\%$ error rate, we obtained a mean improvement in logical errors of $3.3\%$ for a code distance of $5$ and $6.5\%$ for a code distance of $7$. A comparison of decoders with different architectures for a fixed code distance of $5$ indicated the reproducibility and effectiveness of our reoptimization method. In particular, our method had the largest effect on the transformer decoder. We also demonstrated that our approach effectively reduced the number of data needed to train decoders to a certain accuracy. Our method was effective on decoders with a biased noise model of amplitude-phase errors, suggesting its usefulness in a wide range of noise models. It proved feasible with a  continuous function approximation model at a small additional cost. We expect that our method has the possibility of working with a stabilizer code decoder in addition to deep architecture with toric code, which would further contribute to the improvement of QEC.

\end{document}